\pdfoutput=1

\documentclass[11pt]{article}

\usepackage[]{emnlp2021}

\usepackage{times}
\usepackage{latexsym}

\usepackage[T1]{fontenc}

\usepackage[utf8]{inputenc}

\usepackage{microtype}

\usepackage{textgreek}
\usepackage{booktabs,tabularx,multirow}
\usepackage{todonotes}
\usepackage[capitalise,noabbrev]{cleveref}
\usepackage{svg}
\usepackage{colortbl}

\def\parcite#1{\citep{#1}} 
\def\perscite#1{\citet{#1}} 
\def\inparcite#1{\citealp{#1}} 

\usepackage{totcount}
\newtotcounter{todocounter}

\long\def\XXXX#1{}

\usepackage{dingbat}
\def\yesmark{\checkmark}
\def\nomark{\textbf{---}}

\def\rotcell#1{\scriptsize{\rotatebox[origin=c]{90}{\shortstack{#1}}}}
\def\rotmulticell#1#2{\parbox[t]{2mm}{\multirow{#1}{*}{\rotatebox[origin=c]{90}{#2}}}}
\def\multicell#1{\begin{tabular}{@{}c@{}}#1\end{tabular}}
\newcolumntype{H}{>{\setbox0=\hbox\bgroup}c<{\egroup}@{}} 

\usepackage{minibox}
\usepackage{enumitem}
\usepackage{rotating}

\usepackage{float}

\def\countoflanguages{101}
\def\countofpairs{232}

\def\totalsystems{4380}

\def\totalannotationsMillions{2.3 M}
\def\countoflanguagesShowPairsOver{15}
\def\avgTestsetLength{1017}
\def\bleumismatchcount{203}
\def\bleumismatchmedian{1.3}
\def\totalcomparisons{3347}
\def\totalcomparisons0.05{1717}
\def\dependentLanguageCount{17}
\def\dependentCampaigns{333}

\def\dependentTotalEvaluationCountRounded{530000}

\title{To Ship or Not to Ship: \\ An Extensive Evaluation of Automatic Metrics for Machine Translation}

\author{
\minibox[c]{Tom\\Kocmi} \qquad 
\minibox[c]{Christian\\Federmann} \qquad 
\minibox[c]{Roman\\Grundkiewicz} \qquad 
\minibox[c]{Marcin\\Junczys-Dowmunt} \qquad 
\minibox[c]{Hitokazu\\Matsushita} \qquad 
\minibox[c]{Arul\\Menezes}\\[4mm]
  Microsoft\\ 1 Microsoft Way\\ Redmond, WA 98052, USA \\
  \texttt{\{tomkocmi,chrife,rogrundk,marcinjd,himatsus,arulm\}@microsoft.com} \\}
  
\date{}

\begin{document}

\maketitle

\begin{abstract}

Automatic metrics are commonly used as the exclusive tool for declaring the superiority of one machine translation system's quality over another. The community choice of automatic metric guides research directions and industrial developments by deciding which models are deemed better. 
Evaluating metrics correlations with sets of human judgements has been limited by the size of these sets. 
In this paper, we corroborate how reliable metrics are in contrast to human judgements on -- to the best of our knowledge -- the largest collection of judgements reported in the literature. Arguably, pairwise rankings of two systems are the most common evaluation tasks in research or deployment scenarios. Taking human judgement as a gold standard, we investigate which metrics have the highest accuracy in predicting translation quality rankings for such system pairs. 
Furthermore, we evaluate the performance of various metrics across different language pairs and domains. Lastly, we show that the sole use of BLEU impeded the development of improved models leading to bad deployment decisions.
We release the collection of \totalannotationsMillions{} sentence-level human judgements for \totalsystems{} systems for further analysis and replication of our work.

\end{abstract}

\section{Introduction}

Automatic evaluation metrics are commonly used as the main tool for comparing the translation quality of a pair of machine translation (MT) systems \parcite{marie2021scientific}. The decision of which of the two systems is better is often done without the help of human quality evaluation which can be expensive and time-consuming. However, as we confirm in this paper, metrics badly approximate human judgement \cite{wmt2020metrics}, can be affected by specific phenomena \cite{zhang2019translationese,graham-etal-2020-statistical,mathur2020tangled, freitag2021experts} or ignore the severity of translation errors \cite{freitag2021experts}, and thus may mislead system development by incorrect judgements. Therefore, it is important to study the reliability of automatic metrics and follow best practices for the automatic evaluation of systems.


Significant research effort has been applied to evaluate automatic metrics in the past decade, including annual metrics evaluation at the WMT conference and other studies \parcite{wmt2007metrics, przybocki2009nist, wmt2015metrics, wmt2020metrics}. Most research has focused on comparing sentence-level (also known as segment-level) correlations between metric scores and human judgements; or system-level (e.g., scoring an entire test set) correlations of individual system scores with human judgement. \perscite{mathur2020tangled} emphasize that this scenario is not identical to the common use of metrics, where instead, researchers and practitioners use automatic scores to compare a pair of systems, for example when claiming a new state-of-the-art, evaluating different model architectures, deciding whether to publish results or to deploy new production systems. 

The main objective of this study is to find an automatic metric that is best suited for making a \emph{pairwise ranking of systems} and measure how much we can rely on the metric's binary verdicts that one MT system is better than the other.
We design a new methodology for pairwise system-level evaluation of metrics and use it on -- to the best of our knowledge -- the largest collection of human judgement of machine translation outputs which we release publicly with this research.
We investigate the reliability of metrics across different language pairs, text domains and how statistical tests over automatic metrics can help to increase decision confidence. We examine how the common use of BLEU over the past years has possibly negatively affected research decisions. Lastly, we re-evaluate past findings and put them in perspective with our work.
This research evaluates not only the utility of MT metrics in making pairwise comparisons specifically -- it also contributes to the general assessment of MT metrics.

Based on our findings, we suggest the following best practices for the use of automatic metrics:

\begin{enumerate}[noitemsep]
  \item Use a pretrained metric as the main automatic metric; we recommend COMET. Use a string-based metric for unsupported languages and as a secondary metric, for instance ChrF. Do not use BLEU, it is inferior to other metrics, and it has been overused.
  \item Run a paired significance test to reduce metric misjudgement by random sampling variation.
  \item Publish your system outputs on public test sets to allow comparison and recalculation of different metric scores.
\end{enumerate}

\section{Data}

In this section, we describe test sets, the process for collecting human assessments, and MT systems used in our analysis.
We publish all human judgements, metadata, calculated metrics scores, and the code with replication of our findings and promoting further research. We cannot release the proprietary test sets and so system outputs for legal reasons. The collection is available at \url{https://github.com/MicrosoftTranslator/ToShipOrNotToShip}. Moreover, we plan to evaluate new metrics emerging in the future.

\subsection{Test sets}
\label{ssec:testsets}

When evaluating our models, we use internal test sets where references are translated by professional translators from monolingual data. \perscite{freitag2020bleu_might_be_guilty} have demonstrated that the quality of test set references plays an important role in automatic metric quality and correlation with human judgement.
To maintain a high quality of our test sets, we create them by a two-step translation process: the first professional translator translates the text manually without post-editing followed by a second independent translator confirming the quality of the translations. The human translators are asked to translate sentences in isolation; however, they see context from other sentences.

The test sets are created from authentic source sentences, mostly drawn from news articles (news domain) or cleaned transcripts of parliamentary discussions (discussion domain). The news domain test sets are used in both directions, where the authentic side is mostly English, Chinese, French, or German. The discussion domain test sets are used in the direction from authentic source to translationese reference, e.g., we have two distinct test sets, one for English to Polish and second for Polish to English. Furthermore, some systems are evaluated using various other test sets. 

We evaluate \countoflanguages{} different languages within \countofpairs{} translation directions.\footnote{We compare metrics only over the intersection of languages supported by all evaluated metrics, which means that we use only 39 different target languages when Prism is part of the evaluation.} The size of the test sets can vary, and more than one test set or its subsets can be used for a single language direction. The average size of our test sets is \avgTestsetLength{} sentences. The distribution of evaluated systems is not uniform, some language pairs are evaluated only a few times and while others repeatedly with different systems. The majority of the language pairs are English-centric, however, we evaluate a small set of French, German, and Chinese-centric systems (together only 90 system pairs). Details about the system counts of evaluated language pairs and average test set sizes can be found in the Appendix in \cref{tab:lang_pairs_counts}.

\subsection{Manual quality assessment}
\label{ssec:human_judgement}

Our human evaluation is run periodically to confirm translation quality improvements by human judgements. For this analysis, we use human annotations performed from the middle of 2018 until early 2021. All human judgements were collected with identical settings with the same pool of human annotators. Thus, the human annotations should have similar distributions and characteristics. 

The base unit of our human evaluation is called \emph{a campaign}, in which we commonly compare two to four systems in equal conditions: We randomly draw around 500 sentences from a test set, translate them with each system and send them to human assessment. Each human annotator on average annotates 200 sentences, thus a system pair is evaluated by five different annotators (each annotating distinct set of sentences translated by both systems).

We use source-based Direct Assessment (\emph{DA}, \inparcite{graham-etal-2013-continuous}) for collecting human judgements, where bilingual annotators are asked to rate all translated sentences on a continuous scale between 0 to 100 against source sentence without access to reference translations. This eliminates reference bias from human judgement by design.

We use the implementation of DA in the Appraise evaluation framework \parcite{federmann-2018-appraise}, the same as is used in WMT since 2016 for out-of-English human evaluation \parcite{bojar2016wmt}.

We do not use crowd workers as human annotators. Instead, we use paid bi-lingual speakers that are familiar with the topic and well-qualified in the annotation process. Moreover, we track their performance, and those who fail quality control \parcite{graham-etal-2013-continuous} are permanently removed from the pool of annotators, so are their latest annotations.
This increases the overall quality of our human quality assessment. 

We have two additional constraints in contrast to the original DA. Firstly, each system is compared on the same set of sentences which removes the problem of a system potentially benefitting from an easier set of randomly selected sentences. Moreover, it allows us to use a stronger paired test that compares differences in scoring of equal sentences instead of an unpaired one that evaluates scores of both systems in isolation. We use the Wilcoxon signed-rank test \parcite{wilcoxon1946individual} in contrast to the Mann-Whitney U-test \parcite{mann1947test} originally suggested for DA \parcite{graham2017canmachine}. 
Secondly, each annotator is assigned the same number of sentences for each evaluated system which mitigates bias from different rating strategies as each system is affected evenly by each annotator.

When calculating the system score, we take the average of human judgements.\footnote{We do not assume a normal distribution of annotator's annotations; therefore, we do not use z-score transformation.}
We analyze human judgements for \totalsystems{} systems and \totalannotationsMillions{} annotated sentences. This data is one and a half orders of magnitude larger than the data used at WMT Metric Shared Tasks, which evaluate around 170 systems each year (see \cref{sec:metaanalysis}).

\subsection{Systems}
\label{ssec:systems}

We evaluate competing systems against human judgement. The system pairs could be separated into three groups: (1) model improvements, (2) state-of-the-art evaluation, and (3) comparisons with third-party models. The first group contains system pairs where one system is a strong baseline (usually our highest quality system so far) and the second system is an improved candidate model; this group evaluates stand-alone models without additional pre- and post-processing steps (e.g., rule-based named entity matching). The second group contains pairs of the candidate for the new best performing system and the current best performing system. The third group compares our best-performing model at the time with a publicly available third-party MT system.

Analyzing the variety of systems, hyperparameters, training data, and even architectures is out of the scope of this paper. However, all models are based on neural architectures.

\begin{figure*}[h]
\begin{center}
\scalebox{0.8}{
    \includegraphics{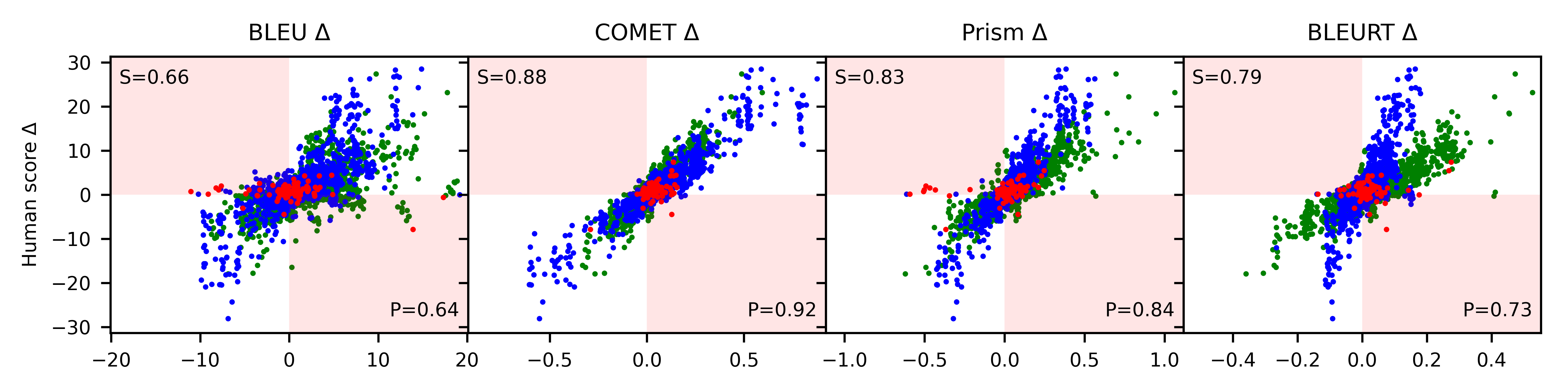}
    }
\end{center}
\caption{\XXXX{switch to SVG}Each point represents a difference in average human judgement (y-axis) and a difference in automatic metric (x-axis) over a pair of systems. Blue points are system pairs translating from English; green points into English; red points are non-English system pairs (a few French, German, or Chinese-centric system pairs). We report Spearman's \textrho{} correlation in the top left corner and Pearson's r in the bottom right corner. Metrics disagree with human ranking for system pairs in pink quadrants. Other metrics are in \cref{app:metric_deltas} in the Appendix.}
\label{fig:deltas_comparison_excerpt}
\end{figure*}

\section{Automatic metrics}
\label{sec:automatic_metrics}

\begin{table}[t]
\centering\small
\begin{tabular}{clcccccHH}
\toprule
 & Metric & \rotcell{Sentence-level} & \rotcell{Use human data} & \rotcell{Need reference} & \rotcell{Multiple refer.} & \rotcell{Languages} & \rotcell{Citation} & \rotcell{Implementation note}\\
\midrule
\rotmulticell{5}{string-based}
 & BLEU         & \nomark{}  & \nomark{}  & \yesmark{} & \yesmark{} & any & \perscite{papineni2002bleu} & SacreBLEU 1.5.0 \parcite{post2018sacrebleu}\\
  & CharacTER         & \yesmark{} & \nomark{}  & \yesmark{} & \nomark{} & any & \perscite{wang2016character} &  \\
 & ChrF         & \yesmark{} & \nomark{}  & \yesmark{} & \nomark{} & any & \perscite{popovic2015chrf} & SacreBLEU 1.5.0 \parcite{post2018sacrebleu}\\
 & EED          & \yesmark{} & \nomark{}  & \yesmark{} & \nomark{} & any & \perscite{stanchev2019eed} &  \\
 & TER          & \yesmark{} & \nomark{}  & \yesmark{} & \nomark{} & any & \perscite{snover2006study} & SacreBLEU 1.5.0 \parcite{post2018sacrebleu}\\
\midrule
\rotmulticell{7}{pretrained}
 & BERTScore   & \yesmark{} & \nomark{}  & \yesmark{} & \nomark{} & 104 & \perscite{zhang2020bertscore} & \\
 & BLEURT       & \yesmark{} & \yesmark{} & \yesmark{} & \nomark{} & * & \perscite{sellam2020bleurt} & BLEURT base\\
 & COMET        & \yesmark{} & \yesmark{} & \yesmark{} & \nomark{} & 100 & \perscite{rei2020comet} & COMET large DA\\
 & ESIM        & \yesmark{} & \yesmark{} & \yesmark{} & \nomark{} & 104 & \perscite{mathur2019putting} & \\
 & Prism        & \yesmark{} & \nomark{}  & \yesmark{} & \nomark{} & 39  & \perscite{thompson2020prism} &  \\
 & COMET-src & \yesmark{} & \yesmark{} & \nomark{}  & n/a    & 100 & \perscite{rei2020comet} & COMET large QE\\
 & Prism-src & \yesmark{} & \nomark{}  & \nomark{}  & n/a    & 39  & \perscite{thompson2020prism} & \\
\bottomrule
\end{tabular}
\caption{Comparison of selected string-based and pretrained automatic evaluation metrics. We mark metrics designed to work at sentence-level, fine-tuned on human judgements, requiring reference(s), or supporting multiple references, and report the number of supported languages. *BLEURT is built on top of English-only BERT \parcite{devlin2019bert} in contrast to BERTScore and ESIM that use multilingual BERT.}
\label{tab:metrics_description}
\end{table}

In this study, we investigate metrics that were shown to provide promising performance in recent studies (see \cref{sec:metaanalysis}) and currently most widely used metrics in the MT field.\footnote{The YiSi -- high correlating metric \parcite{wmt2019metrics} -- was not publicly available at the time of our evaluation.} We focus on language-agnostic metrics, therefore we do not include metrics supporting only a small set of languages. The full list of evaluated metrics and their main features is presented in \cref{tab:metrics_description}.

Two categories of automatic machine translation metrics can be distinguished: (1) string-based metrics and (2) metrics using pretrained models. The former compares the coverage of various substrings between the human reference and MT output texts. String-based methods largely depend on the quality of reference translations. However, their advantage is that their performance is predictable as it can be easily diagnosed which substrings affect the score the most. The latter category of pretrained methods consists of metrics that use pretrained neural models to evaluate the quality of MT output texts given the source sentence, the human reference, or both. They are not strictly dependent on the translation quality of the human reference (for example, they can better evaluate synonyms or paraphrases). However, their performance is influenced by the data on which they have been trained. Moreover, the pretrained models introduce a black-box problem where it is difficult to diagnose potential unexpected behavior of the metric, such as various biases learned from training data.

For all metrics, we use the recommended implementation. See \cref{app:implementation_details} for implementation details.
Most metrics aim to achieve a positive correlation with human assessments, but some error metrics, such as TER, aim for a negative correlation. We simply negate scores of metrics with anticipated negative correlations.
Pretrained metrics usually do not support all languages, therefore to ensure comparability, we evaluate metrics on a set of language pairs supported by all metrics.

\section{Evaluation}

\subsection{Pairwise score differences}
\label{ssec:metric_delta}

Most previous works studied the system-level evaluation of MT metrics in an isolated scenario correlating individual systems with human judgements \parcite{wmt2007metrics, wmt2020metrics}. They have mostly employed Pearson's correlation (see \cref{sec:metaanalysis}) as suggested by \perscite{wmt2014metrics} and evaluated each language direction separately. However, \perscite{mathur2020tangled} suggest using a pairwise comparison as a more accurate scenario for the general use of metrics. 

As the primary unit, we use the difference in metric (or human) scores between system A and B:
\[\Delta = \mbox{score}(\mbox{System A})-\mbox{score}(\mbox{System B})\]
We gather all system pairs from each campaign separately as only systems within a campaign are evaluated under equal conditions. All campaigns compare two, three, or four systems, which results in one, three, or six system pairs, respectively. 

To understand the relationship between metrics and absolute human differences, we plot these differences and calculate Pearson's and Spearman's correlations in \cref{fig:deltas_comparison_excerpt}. All metrics exhibit a positive correlation with human judgements but differ in behavior. For example, COMET has the smallest deviation which results in the highest correlation with human judgements. However, when we evaluate into-English and from-English language directions separately, we observe that COMET, Prism, and mainly BLEURT have inconsistent value ranges for different language pairs.\footnote{A possible explanation for BLEURT is that it is trained on English-only. But this does not explain other metrics.} 

Hence, we cannot assume equal scales for one metric and different language pairs, so we can not use Pearson's nor Spearman's correlation in pairwise metrics evaluation. Nonetheless, we provide both correlations in Appendix \cref{tab:system_level_description_appendix} for the complete picture.


\subsection{Pairwise system-level metric quality}

As standard correlation cannot be used, we investigate a different approach to evaluation. We advocate that the most important aspect of a metric is to make reliable binary pairwise decisions (i.e., which of two systems provides a higher translation quality) without the focus on the magnitude of difference.\footnote{The value of score difference (e.g., a difference of 2 BLEU) is important mainly to measure the confidence of a ranking decision.} Therefore, given the size of our data set, we propose to use accuracy on binary comparisons: which system is better when human rankings are considered gold labels.

We define the accuracy as follows. For each system pair, we calculate the difference of the metric scores (metric$\Delta$) and the difference in average human judgements (human$\Delta$). We calculate accuracy for a given metric as the number of rank agreements between metric and human deltas divided by the total number of comparisons:
\[\mbox{Accuracy} = \frac{|\mbox{sign}(\mbox{metric} \Delta)~=~\mbox{sign}(\mbox{human} \Delta)|}{|\mbox{all system pairs}|}\]
Assuming human judgements as a gold labels, accuracy gets an intrinsic meaning of how ,,reliable'' a given metric is when making pairwise comparisons. On the other hand, accuracy does not take into account that two systems can have comparable quality, and thus the accuracy of a metric can be over-estimated by chance if a small human score difference has the same sign as the difference in a metric score.
To overcome this issue, we also calculate accuracy over a subset of system pairs, where we remove system pairs that are deemed to not be different based on Wilcoxon's signed-rank test over human judgements. 

In order to estimate the confidence interval for accuracy, we use the bootstrap method \parcite{efron1994introduction}, for more details see \cref{app:accuracy_confidence_interval}. We consider all metrics that fall into the 95\% confidence interval of the best performing metric to be comparable. We visualize the clusters of best-performing metrics in our analysis with a grey background of table cells.

\section{Results}

\subsection{Which metric is best suited for pairwise comparison?}
\label{ssec:best_metric}

\begin{table}[t]
\centering
\scalebox{0.85}{
\begin{tabular}{lrrrr|r}
\toprule
{} &                                All &                               0.05 &                               0.01 &                              0.001  & Within\\
n         &                               3344 &                               1717 &                               1420 &                               1176 &                                541 \\
\midrule
COMET     &  \cellcolor{black!15}\textbf{83.4} &  \cellcolor{black!15}\textbf{96.5} &  \cellcolor{black!15}\textbf{98.7} &  \cellcolor{black!15}\textbf{99.2} &  \cellcolor{black!15}\textbf{90.6} \\
COMET-src &           \cellcolor{black!15}83.2 &                               95.3 &                               97.4 &                               98.1 &           \cellcolor{black!15}89.1 \\
Prism     &                               80.6 &                               94.5 &                               97.0 &                               98.3 &                               86.3 \\
BLEURT    &                               80.0 &                               93.8 &                               95.6 &                               98.2 &                               84.1 \\
ESIM      &                               78.7 &                               92.9 &                               95.6 &                               97.5 &                               82.8 \\
BERTScore &                               78.3 &                               92.2 &                               95.2 &                               97.4 &                               81.0 \\
ChrF      &                               75.6 &                               89.5 &                               93.5 &                               96.2 &                               75.0 \\
TER       &                               75.6 &                               89.2 &                               93.0 &                               96.2 &                               73.9 \\
CharacTER &                               74.9 &                               88.6 &                               91.9 &                               95.2 &                               74.1 \\
BLEU      &                               74.6 &                               88.2 &                               91.7 &                               94.6 &                               74.3 \\
Prism-src &                               73.4 &                               85.3 &                               87.6 &                               88.9 &                               77.4 \\
EED       &                               68.8 &                               79.4 &                               82.4 &                               84.6 &                               68.2 \\
\bottomrule
\end{tabular}
}
\caption{Accuracies for binary comparisons for ranking system pairs. Column ``All'' shows the results for system pairs. Each following column evaluates accuracy over a subset of systems that are deemed different based on human judgement and a given alpha level in Wilcoxon's test. Column ``Within'' represents a subset of systems where the human judgement p-value is between 0.05 and 0.001.
``n'' represents the number of system pairs used to calculate accuracies in a given column. Only the scores in each column are comparable. Results with a grey background are considered to be tied with the best metric.}
\label{tab:system_level_description}
\end{table}

\begin{table*}[t]
\centering
\scalebox{0.95}{
\begin{tabular}{l|l|ll|ll|ll}
\toprule
{} &                         Everything &                            Into EN &                            From EN &                          Non Latin &                          Logograms &                            Non WMT &                             Discussion \\

n         &                               1717 $\downarrow$ &                                922 &                                768 &                                131 &                                 44 &                                484 &                                 78 \\
\midrule
COMET     &  \cellcolor{black!15}\textbf{96.5} &  \cellcolor{black!15}\textbf{95.3} &  \cellcolor{black!15}\textbf{98.3} &  \cellcolor{black!15}\textbf{96.2} &  \cellcolor{black!15}\textbf{90.9} &  \cellcolor{black!15}\textbf{97.3} &  \cellcolor{black!15}\textbf{93.6} \\
COMET-src &                               95.3 &                               93.5 &           \cellcolor{black!15}97.7 &           \cellcolor{black!15}95.4 &           \cellcolor{black!15}88.6 &           \cellcolor{black!15}96.7 &  \cellcolor{black!15}\textbf{93.6} \\
Prism     &                               94.5 &                               92.2 &           \cellcolor{black!15}98.2 &  \cellcolor{black!15}\textbf{96.2} &  \cellcolor{black!15}\textbf{90.9} &           \cellcolor{black!15}96.9 &                               83.3 \\
BLEURT    &                               93.8 &                               93.8 &                               95.1 &           \cellcolor{black!15}93.1 &           \cellcolor{black!15}84.1 &                               94.6 &           \cellcolor{black!15}89.7 \\
ESIM      &                               92.9 &                               90.6 &                               96.6 &           \cellcolor{black!15}93.9 &           \cellcolor{black!15}86.4 &                               94.8 &                               76.9 \\
BERTScore &                               92.2 &                               91.2 &                               94.1 &           \cellcolor{black!15}95.4 &           \cellcolor{black!15}88.6 &                               92.8 &                               71.8 \\
ChrF      &                               89.5 &                               88.7 &                               91.0 &           \cellcolor{black!15}95.4 &           \cellcolor{black!15}88.6 &                               89.7 &                               57.7 \\
TER       &                               89.2 &                               87.6 &                               91.7 &                               90.1 &                               72.7 &                               90.9 &                               70.5 \\
CharacTER &                               88.6 &                               86.4 &                               91.7 &                               88.5 &                               70.5 &                               91.9 &                               69.2 \\
BLEU      &                               88.2 &                               86.9 &                               90.5 &           \cellcolor{black!15}92.4 &           \cellcolor{black!15}79.5 &                               89.9 &                               61.5 \\
Prism-src &                               85.3 &                               80.8 &                               91.4 &                               84.0 &                               65.9 &                               91.7 &                               84.6 \\
EED       &                               79.4 &                               75.1 &                               84.8 &                               82.4 &                               54.5 &                               83.1 &                               60.3 \\
\bottomrule
\end{tabular}
}
\caption{Accuracies for ranking system pairs. Each column represents a different subset of significantly different system pairs with alpha level 0.05. Results with a grey background are considered to be tied with the best metric. Accuracies across columns are not comparable as they compare different sets of systems.}
\label{tab:scenarios_accuracies}
\end{table*}

In this section, we examine all available system pairs and investigate which metric is best suited for making a pairwise comparison. 

The results presented in \cref{tab:system_level_description} show that pretrained methods (except for Prism-src) generally have higher accuracy than string-based methods, which confirms findings from other studies \parcite{wmt2018metrics, wmt2019metrics, wmt2020metrics}.
COMET reaches the highest accuracy and therefore is the most suited for ranking system pairs. The runner-up is \emph{COMET-src}, which is a surprising result because, as a quality estimation metric, it does not use a human reference. This opens possibilities to use monolingual data in machine translation systems evaluation in an effective way. On the other hand, the second reference-less method \emph{Prism-src} does not reach high accuracy, struggling mainly with into-English translation directions (see \cref{app:metric_deltas} in the Appendix). 
In terms of string-based metrics, the highest accuracy is achieved by ChrF, which makes it a better choice for comparing system pairs than the widely used BLEU.

To minimize the risk of being affected by random flips due to a small human score delta, we also explore the accuracy after removing systems with comparable performance with respect to Wilcoxon's test over human judgements. We incrementally remove system pairs not significantly different with alpha levels of 0.05, 0.01, and 0.001. As expected, removing pairs of most likely equal-quality systems increases the accuracy, however, no metric reaches 100\% accuracy even for a set of strongly different systems with an alpha level of 0.001. This implies that either current metrics cannot fully replace human evaluation or remaining systems are incorrectly assessed by human annotators.\footnote{An alpha level of 0.001 could (mis)lead to the conclusion that 0.1\% of human judgements are incorrect. However, the alpha level only determines if two systems are different enough and cannot be used to conclude that a human pairwise rank decision is incorrect.}
Moreover, we observe that the ordering of metrics by accuracy remains the same even after removing system pairs with comparable performance, which implies that accuracy is not negatively affected by non-significantly different system pairs. 
Due to that where we analyze only subsets of the data, we use systems that are statistically different by human judgement with an alpha level of 0.05.

\perscite{wmt2019metrics} have observed that system outliers, i.e., systems easily differentiated from other systems, can inflate Pearson's correlation values. Moreso, \perscite{mathur2020tangled} demonstrated that after removing outliers some metrics would actually have negative correlation with humans. To analyze if outliers might affect our accuracy measurements and the ordering of metrics, we analyze a subset of systems with human judgement p-values between 0.05 and 0.001, i.e. removing system pairs that have equal quality and outlier system pairs that are easily distinguished. From column ``Within'' in \cref{tab:system_level_description}, we see that the ordering of metrics remains unchanged. This shows that accuracy is not affected by outliers making it more suitable for metrics evaluation than Pearson's $\rho$.

\subsection{Are metrics reliable for non-English languages and other scenarios?}

The superior performance of pretrained metrics raises the question if unbalanced annotation data might be responsible; around half of the systems translate into English. Moreover, COMET and BLEURT are fine-tuned on human annotations from WMT on the news domain. This could lead to an unfair advantage when being evaluated w.r.t. human judgements.\footnote{We double-checked and removed all campaigns containing test sets from WMT 2015 to 2020 from our work and analysis.}
To shed more light on metrics behavior and robustness, we analyze various subsets, including into and from English translation directions, languages with non-Latin scripts, and non-news domain. 

We showed in \cref{ssec:metric_delta} that some metrics perform differently for systems translating from and into English. Analyzing this scenario in \cref{tab:scenarios_accuracies} reveals that BLEURT does better (the second best metric) for ``into English'' translation compared to other metrics. It is surprising that BLEURT has a high accuracy for unseen ``from English'' pairs which suggests that BLEURT might have learned some kind of string-matching.
We also observe in \cref{tab:scenarios_accuracies} gains for Prism for the ``from English'' directions. The overall ranking of metrics, however, remains similar which confirms that the high accuracy of pretrained methods compared to the string-based ones cannot be attributed to the abundance of system pairs with English as the target.

When investigating language pairs with non-Latin (Arabic, Russian, Chinese, ...) or logogram-based scripts (Chinese, Korean and Japanese) as the target languages, we observe a slight drop in metric ranks for some pretrained metric in contrast to higher score for ChrF. This indicates that non-Latin scripts might be a challenge for pretrained metrics but more analysis would be required here. For an summary on individual language pairs, refer to \cref{tab:individual_language_pairs} in the Appendix.

We also investigate if some pretrained methods might have an unfair advantage due to being fine-tuned on human assessments in the news domain. For this, we analyze a subset of news test sets with target languages that were not part of WMT human evaluation (i.e., languages which those methods have not been fine-tuned on) and call this set ``non-WMT'', and also system pairs evaluated on a proprietary test sets in the EU parliamentary discussions domain covering ten languages. 
Neither results on non-WMT nor discussion domains in \cref{tab:scenarios_accuracies} show a change in the ranking of metrics, suggesting that COMET is not overfitted to the WMT news domain or WMT languages. 
Somewhat surprisingly, we actually see a drop in accuracy for the string-based metrics for the discussion domain. We speculate this might be due to their inability to forgivingly match disfluent utterances to expected fluent translations \parcite{salesky2019fluent}.

Overall, the results for various subsets show a similar ordering of metrics based on their accuracy, confirming the general validity of our results.

\subsection{Are statistical tests on automatic metric worth it?}
\label{ssec:statistical_tests}

\begin{table}
\centering
\begin{tabular}{ll|llHHH}
\toprule
{} &        No test &      Boot. $\downarrow$ &  Type II Err. &                n &           boot n \\
\midrule
COMET     &  \textbf{83.4} &  \textbf{95.1} &  204 (\textbf{17.3\%}) &  \textbf{3344.0} &  \textbf{2093.0} \\
COMET-src &           83.2 &           94.2 &           242 (19.4\%) &  \textbf{3344.0} &           2023.0 \\
BLEURT    &           80.0 &           92.0 &           349 (25.4\%) &  \textbf{3344.0} &           1952.0 \\
Prism     &           80.6 &           91.3 &           200 (18.3\%) &  \textbf{3344.0} &           2231.0 \\
BERTScore &           78.3 &           87.9 &           244 (20.9\%) &  \textbf{3344.0} &           2154.0 \\
ChrF      &           75.6 &           85.4 &           350 (27.3\%) &  \textbf{3344.0} &           2044.0 \\
BLEU      &           74.6 &           83.4 &           378 (27.4\%) &  \textbf{3344.0} &           1942.0 \\
Prism-src &           73.4 &           81.5 &           325 (29.4\%) &  \textbf{3344.0} &           2219.0 \\
\bottomrule
\end{tabular}
\caption{The first column shows accuracy for all system pairs and represent situation, where we would trust any small score difference. The second column shows accuracy, where we ignore systems considered to be tied with respect to the paired bootstrap resampling test. The third column represents the number of system pairs incorrectly decided to be non-significantly different by the paired bootstrap resampling and the percentage from all non-significant systems.}
\label{tab:statitical_tests}
\end{table}

\perscite{mathur2020tangled} studied the effects of statistical testing of automatic metrics and observed that even large metric score differences can disagree with human judgement. They have shown that even for a BLEU delta of 3 to 5 points, a quarter of these systems are judged by humans to differ insignificantly in quality or to contradict the verdict of the metric. In our analysis, we have \bleumismatchcount{} system pairs deemed statistically significant by humans (p-value smaller than 0.05) for which using BLEU results in a flipped ranking compared to humans. The median BLEU difference for these system pairs is \bleumismatchmedian{} BLEU points.
This is concerning as BLEU differences higher than one or two BLEU points are commonly and historically considered to be reliable by the field. 

In this section, we corroborate that statistical significance testing can largely increase the confidence of the MT quality improvement and increase the accuracy of metrics. We compare how accurate a metric would be under two situations: either when not using statistical testing and solely trusting in the metric score difference; or when using statistical testing and throwing away systems that are not statistically different.

We evaluated the first situation in \cref{ssec:best_metric} and the results are equal with the first column of \cref{tab:system_level_description}. 
For the second situation, we calculate accuracy only over the system pairs that are statistically different.
We use paired bootstrap resampling \parcite{koehn2004statistical}, a non-parametric test, to calculate the statistical significance for a pair of systems.\footnote{Approximate randomization \parcite{riezler2005approximate} can be used as an alternative test, and for metrics based on the average of sentence-level scores, we can use also tests such as the Student t-test.}

Additionally, the second situation introduces type II errors which represent systems where the statistical significance test rejected a system pair as being non-significant, but humans would judge the given pair as significantly different. In other words, it shows how many system pairs are incorrectly rejected as non-significantly different. See \cref{app:compare_statistical_tests} for a detailed explanation.

From the results in \cref{tab:statitical_tests}, we can see that if we apply paired bootstrap resampling on automatic metrics with an alpha level 0.05 the accuracy increases by around 10\% for all metrics in contrast to not using statistical testing. On the other hand, when using statistical testing, we introduce type II errors, where 17.3\%, for COMET,\XXXX{check the final number} of non-significantly different system pairs are deemed significantly different by humans.\footnote{Wilcoxon's test on human judgement and alpha level 0.05.} 

In conclusion, we corroborate that using statistical significance tests largely increases reliability in automatic metric decisions. We encourage the usage of statistical significance testing, especially in the light of \perscite{marie2021scientific} who show that statistical significance tests are widely ignored.

\subsection{Does BLEU sabotage progress in MT?}

\perscite{freitag2020bleu_might_be_guilty} have shown that reference translations with string-based metrics may systematically bias against modeling techniques known to improve human-judged quality and raised the question of whether previous research has incorrectly discarded approaches that improved the quality of MT due to the use of such references and BLEU. They argue that the use of BLEU might have mislead many researcher in their decisions.

In this section, we investigate the hypothesis if the usage of BLEU negatively affects model selection. To do so, we compare two groups of system pairs based on the premise if they could be directly affected by BLEU. The first group contains pairs of incremental improvements of our systems. We can assume that incremental models use similar architecture, data, and settings, although we do not study particular changes. We use BLEU as the main automatic metric to guide model development. If BLEU shows improvements, we evaluate models with human judgements to make a final deployment decision. Therefore, systems with degraded BLEU scores which would be deemed improved by humans are missing in this group as we reject them based on BLEU scores during development. 
The second group contains independent system pairs, which use different architectures, data, settings, and therefore BLEU has not been used to preselect them. In this group, we compare our systems with publicly available third-party MT systems.

We compare three models within the same campaign, two internal\footnote{The pair of internal models contains the best model from the last year and our latest improved model.} and one external system. Thus, the same annotators annotated the same sentences from all three systems under the same conditions. We call system pairs comparisons between two internal models ``incremental'', and comparisons between the newer internal model and the external model as ``independent''.

Over the past three years we carried out \dependentCampaigns{} campaigns across \dependentLanguageCount{} language pairs (each campaign comparing three models), resulting in almost \dependentTotalEvaluationCountRounded{} human annotations.

\begin{table}
\centering
\begin{tabular}{lrr}
\toprule
{} &      Incremental &    Independent \\
n         &                                161 $\downarrow$ &                                246 \\
\midrule
BLEU      &  \cellcolor{black!15}\textbf{99.4} &                               90.7 \\
BERTScore &           \cellcolor{black!15}98.8 &                               91.5 \\
ESIM      &           \cellcolor{black!15}98.8 &                               92.3 \\
Prism     &           \cellcolor{black!15}98.1 &                               94.3 \\
ChrF      &           \cellcolor{black!15}98.1 &                               91.5 \\
COMET     &           \cellcolor{black!15}98.1 &  \cellcolor{black!15}\textbf{98.4} \\
COMET-src &           \cellcolor{black!15}97.5 &           \cellcolor{black!15}98.8 \\
CharacTER &           \cellcolor{black!15}97.5 &                               89.8 \\
Prism-src &           \cellcolor{black!15}96.9 &                               92.7 \\
BLEURT    &                               96.9 &                               93.5 \\
TER       &                               95.7 &                               91.5 \\
EED       &                               78.9 &                               78.0 \\
\bottomrule
\end{tabular}
\caption{Evaluation of incremental and independent system pairs. We use a subset of \dependentCampaigns{} system pairs significantly different based on Wilcoxon's test and alpha level of 0.05 over human judgement. Results with grey background are considered tied with the best metric.}
\label{tab:dependent}
\end{table}

The results in \cref{tab:dependent} show that for independent systems, the ranking of the metrics is comparable with results in \cref{tab:scenarios_accuracies}. Pretrained metrics generally outperform string-based ones and COMET is in the lead. However, when inspecting the incremental systems, BLEU wins. This indicates that BLEU influenced our model development and we rejected models that would have been preferred by humans.

Another possible explanation is that systems preselected by BLEU are easy to differentiate by all metrics. This could explain why all metrics have high accuracy in contrast to the ``Independent'' column and most of them are in a single cluster. 

In conclusion, results showing BLEU as the metric with the highest accuracy where we would expect pretrained metrics to dominate, suggests that BLEU affected system development and we rejected improved models due to the erroneous degradation seen in the BLEU score. However, this is indirect evidence as for sound conclusions we would need to evaluate those rejected systems with other metrics and human judgement as well.

\begin{table*}[h]
\centering\small
\begin{tabular}{clHllHlllllll}
\toprule
& WMT Metric task &  & 
\citeyear{wmt2020metrics} &
\citeyear{wmt2020metrics} & 
\begin{tabular}{@{}c@{}}\perscite{mathur2020tangled}\\no outliers\end{tabular} &
\citeyear{wmt2019metrics} & 
\citeyear{wmt2018metrics} & 
\citeyear{wmt2017metrics} & 
\citeyear{wmt2016metrics} & 
\citeyear{wmt2015metrics} & 
\citeyear{wmt2014metrics} & 
\citeyear{wmt2013metrics} \\ 
 & $\downarrow$ n & 2761 & 168 (no outliers) & 184 & 179 & 225 & 149 & 152 & 120 & 121 & 92 & 135 \\
\midrule
\rotmulticell{8}{string-based}
& BLEU      &       & .740 (.727) & .837 (.832) & .825 & .906 & .955 & .910 & .873 & .841 & \textbf{.910} & .845 \\
& CharacTER &       & .735 (.723) & .873 (.871) & .910 & .942 & .964 & .932 & \textbf{.938} &  &  &  \\
& ChrF      &       & .743 (.730) & .743 (.864) & .908 & .948 & .959 & \textbf{.942} & .911 & .908 &  &  \\
& EED       &       & .762 (.750) & .888 (.885) & .913 & .951 &  &  &  &  &  &  \\
& METEOR    &       &  &  &  &  &  &  &  & .900 & .884 & \textbf{.878} \\
& NIST      &       &  &  & .797 & .860 & .970 & .921 & .870 & .854 & .899 & .834 \\
& TER       &       & .609 (.668) & .704 (.763) & .865 & .922 & .953 & .918 & .863 & .837 & .860 & .788 \\
& WER       &       &  &  & .862 & .917 & .934 & .913 & .846 & .829 & .818 & .752 \\
\midrule
\rotmulticell{6}{pretrained}
& BEER      &       &  &  & .899 & .942 & \textbf{.973} & .938 & .925 & \textbf{.942} &  &  \\
& BLEURT    &       & .764 (.752) & .902 (.900) &  &  &  &  &  &  &  &  \\
& COMET     &       & .711 (\textbf{.762}) & .853 (\textbf{.908}) &  &  &  &  &  &  &  &  \\
& ESIM      &       & \textbf{.770} (.755) & \textbf{.906} (.902) &  &  &  &  &  &  &  &  \\
& Prism     &       & .677 (.710) & .846 (.886) &  &  &  &  &  &  &  &  \\
& YiSi-1    &       & .759 (.744) & .894 (.890) & \textbf{.942} & \textbf{.967} & \textbf{.973} &  &  &  &  &  \\
\bottomrule
\end{tabular}
\caption{``n'' is sum of systems in each study used to calculate aggregated correlation. The results in brackets are without systems on English into Chinese. Correlations are comparable only within columns.}
\label{tab:metaanalysis}
\end{table*}

\section{Meta Analysis}
\label{sec:metaanalysis}

We analyze findings from past research to put our results in the broader context. We focus on the results on the system-level evaluation, however, a large part of the research studied a sentence-level evaluation. The largest source of metrics evaluation is yearly WMT Metric Shared Task occurring over more than the past ten years \parcite{wmt2007metrics}, where various methods are evaluated with human judgement over the set of submitted systems and language pairs in WMT News Translation Shared Tasks. Recently, \perscite{freitag2021experts} reevaluated two translation directions from WMT 2020 with the multidimensional quality metric framework and raised a concern that general crowd-sourced annotators used in into-English evaluation in WMT prefer literal translations and have a lower quality than some automatic metrics.

Past studies evaluate system-level correlations with Pearson's correlation calculated for each translation direction separately. We are interested in how metrics correlate with human judgement in general across different language pairs. Thus, to generalize the past findings, we use the Hunter-Schmidt method \parcite{hunter2004methods}, which allows combining already calculated correlations with various sizes. We use it to generalize correlations within each study across all language pairs. For this purpose, Hunter-Schmidt is effectively a weighted mean of the raw correlation coefficients. 

Although past studies evaluated a larger number of methods and their variants, we have selected a subset of metrics that are evaluated in more than one study or showed promising performance over other metrics in a given study. When a study evaluated several variants of a metric with various parameters, we selected the setting closest to either the recommended setting in the recent years, such as SacreBLEU, or a setting that is used in the later evaluation study, mainly in \perscite{wmt2020metrics}. 

Meta-analysis in \cref{tab:metaanalysis} shows that pretrained methods outperform string-based methods as concluded by \perscite{wmt2020metrics, wmt2019metrics, wmt2018metrics}. The second important observation is that there was not a single year where BLEU had a higher correlation than ChrF. This supports our conclusions and shows that the MT community had results supporting the deprecation of BLEU as a standard metric for several years. Comparing the pretrained methods, ESIM is the best performing method in general \parcite{wmt2020metrics}, while COMET is the best performing method when removing the suspicious system.

In the study by \perscite{wmt2020metrics}, COMET under-performed other pretrained metrics. We found out that submitted COMET scores failed to score one English-Chinese system with tokenized output. However, we obtain valid COMET scores on that system output when replicating the results. Moreover, we have not seen any problems with COMET on Chinese. As this one system largely skews Pearson's correlation, we also present analysis without English-Chinese systems in \cref{tab:metaanalysis}. 

\section{Discussion}

We corroborate results from past studies that pretrained methods are superior to string-based ones. However, pretrained methods are relatively new techniques and we can potentially discover significant drawbacks, for example, they could resemble biases from training data, fail on particular domains, or prefer fluency over adequacy. Another problem could arise if an MT system would be trained on the same data as the metric was or if it incorporates the same pretrained model, for example, XLM-R \parcite{conneau2020XLM} used by COMET. Pretrained methods support only a selected set of languages and the quality can differ for each of them. Thus, we argue that the string-based method should be used as a secondary metric.

An interesting solution to dissipate potential drawbacks of any metric would be if different research groups preselect a different primary pretrained metric in advance to lead their research decisions and to discover improvements not apparent under other metrics. However, we fear that it could lead to ``metric-hacking'', i.e., picking a metric that confirms results. 
Therefore, we recommend using COMET as the primary metric. And to use ChrF, the best performing string-based method, as a secondary metric and for unsupported languages.

A surprising results is the high accuracy of COMET-src, a reference-free metric. It allows automatic evaluation over monolingual domain-specific testsets as suggested by \perscite{agrawal2021assessing}.

Limitations of BLEU are well-known \parcite{reiter2018structured, mathur2020tangled}.  \perscite{callison2006reevaluating} argued that MT community is overly reliant on it, which \perscite{marie2021scientific} confirmed by showing that 98.8\% of MT papers use BLEU. We present indirect evidence that the over-use of BLEU negatively affects MT development and support deprecation of BLEU as the evaluation standard.

We show that the reliability of metrics decisions can be increased with statistical significance tests. However, \perscite{dror2018hitchhiker} point out the assumption of statistical significance tests that data samples are independent and adequately distributed is rarely true. 
Also, statistical significance tests do not account for random seed variation across training runs. Thus, one should be cautious when making conclusions based on small metrics improvements. \perscite{wasserstein2019moving} give recommendations for a better use of statistical significance testing.

\perscite{marie2021scientific} have shown that almost 40\% of MT papers from 2020 copied score from different papers without recalculating them, which is a concerning trend. Also, new and better metrics will emerge and there is no need to permanently adhering to a single metric. Instead, the simplest and most effective solution to avoid the need to copy scores or stick to obsolete metric is to always publish translated outputs of test sets along with the paper. This allows anyone to recalculate scores with different tools and/or metrics and makes comparisons with past (and future) research easier.

There are some shortcomings in our analysis. We have only a handful of non-English systems, therefore we cannot conclude anything about the behaviour of the metrics for language pairs without English. Similarly, the majority of our language pairs are high-resource, therefore, we cannot conclude the reliability of metrics for low-resource languages. Lastly, many of our translation directions are from translationese into authentic, which as \perscite{zhang2019translationese} showed is the easier direction for systems to score high by human judgement. These are potential directions of future work.

Lastly, we assume that human judgement is the gold standard. However, we need to keep in mind that there can be potential drawbacks of the method used for human judgement or human annotators fail to capture true assessment as \perscite{freitag2021experts} observe. For example, humans cannot explicitly mark critical errors in DA and instead they usually assign low assessment scores.

\section{Conclusion}

We show that metrics can use a different scale for different languages, so Pearson's correlation cannot be used. We introduce accuracy as a novel evaluation of metrics in a pairwise system comparison. 

We use and release a large collection of the human judgement confirming that pretrained metrics are superior to string-based. COMET is the best performing metric in our study, and ChrF is the best performing string-based method. The surprising effectiveness of COMET-src could allow the use of large monolingual test sets for quality estimation.

We do not see any drawbacks of the metrics when investigating various languages or domains, especially, for methods pretrained on human judgement. We present indirect evidence that the over-use of BLEU negatively affects MT development.

We show that statistical testing of automatic metrics largely increases the reliability of a pairwise decision based on automatic metric scores.

We endorse the recommendation for publishing translated outputs of research systems to allow comparisons and recalculation of scores in the future.

\section*{Acknowledgments}
We are grateful for a feedback and review of the paper to many researchers, namely: 
Shuoyang Ding, 
Markus Freitag, 
Hieu Hoang, 
Alon Lavie, 
Jindřich Libovick\'{y}, 
Nitika Mathur, 
Mathias M\"{u}ller, 
Martin Nejedl\'{y}, 
Martin Popel, 
Matt Post, 
Qingsong Ma,
Richardo Rei, 
Thibault Sellam, 
Ale\v{s} Tamchyna, anonymous reviewers, and our colleagues.

\bibliographystyle{acl_natbib}
\bibliography{emnlp2021}

\appendix

\section{Metrics Implementation Details}
\label{app:implementation_details}

We use the most common implementation with default or recommended parameters to simulate standard metric usage. 

For \emph{BLEU} \parcite{papineni2002bleu}, \emph{ChrF} \parcite{popovic2015chrf} and \emph{TER} \parcite{snover2006study} metrics, we use SacreBLEU implementation \url{https://github.com/mjpost/sacrebleu/} version 1.5.0. We use ``mteval-v13a'' tokenizer for all language pairs except for Chinese and Japanese which use their own tokenizer, as is recommended.

For \emph{CharacTER} \parcite{wang2016character}, we use \url{https://github.com/rwth-i6/CharacTER} commit c4b25cb.

For \emph{EED} \parcite{stanchev2019eed}, we use \url{https://github.com/rwth-i6/ExtendedEditDistance} commit f944adc.

For \emph{BERTScore} \parcite{zhang2020bertscore}, we use \url{https://github.com/Tiiiger/bert_score} version 0.3.7.

For \emph{BLEURT} \parcite{sellam2020bleurt}, we use the recommended model ``bleurt-base-128'' and implementation \url{https://github.com/google-research/bleurt} version 0.0.1. It is important to mention, that BLEURT is fine-tuned for English only. Additionally, we evaluated other variants and ``bleurt-large-512'' performed better than recommended variant. We add it in \cref{tab:system_level_description_appendix}.

For \emph{COMET} \parcite{rei2020comet}, we use recommended model ``wmt-large-da-estimator-1719'' and for \emph{COMET-src} we use ``wmt-large-qe-estimator-1719''. The implementation is \url{https://github.com/Unbabel/COMET} in version 0.0.6. We evaluated all other COMET models, but neither performed better than recommended model.

For \emph{Prism} and \emph{Prism-src} \parcite{thompson2020prism}, we use \url{https://github.com/thompsonb/prism} commit 06f10da.

For \emph{ESIM} \parcite{mathur2019putting}, we use \url{https://github.com/nitikam/mteval-in-context}.

\section{Confidence Interval for Metric Accuracy}
\label{app:accuracy_confidence_interval}

To estimate the confidence interval for the best performing metric, we use the bootstrap method \parcite{efron1994introduction}. It creates multiple resamples (with replacement) from a set of observations and calculates accuracy on each of these resamples. 
We employ modified paired bootstrap resampling \parcite{koehn2004statistical}, a method which we also use for testing statistical significance of the metric difference in \cref{ssec:statistical_tests}. However, the usage is different.

To calculate the bootstrap resampling. First, we note the best performing metric on all system pairs from the collection as metric $\alpha$.
We create 10 000 resamples by drawing system pairs with replacements from the collection of all. 
For each resample, we calculate accuracy for all metrics. We note which metrics have equal or higher accuracy than metric $\alpha$ in a given resample. 

If metric $\alpha$ outperforms metric X by less than 95\% of the time, we draw the conclusion that metric X performs on par with 95\% statistical significance to the winning metric $\alpha$.

\section{Comparing Statistical Tests}
\label{app:compare_statistical_tests}

The problem if two systems have the same MT quality is still an open question. Applying statistical tests over the metric scores allows us to confirm if the difference in score is significant or due to a random change based on the set of translated sentences and a given alpha level. To get the gold truth about system equivalence, we employ Wilcoxon's test on human judgement and alpha level 0.05. We use paired bootstrap resampling approach as the statistical test for automatic metrics. Unfortunately, we cannot directly compare the outputs of two statistical tests (for example, the Wilcoxon test on human judgements with the bootstrap resampling on metric scores) as even with the same alpha level, these tests have a different power. Therefore, we need to investigate it in isolation.

The null hypothesis in our setting is that both evaluated systems have the same translation quality. There are two possible outcomes of a statistical test: accept the null hypothesis (i.e. MT quality of systems is \emph{not significantly} different) or reject the null hypothesis (i.e. MT quality of systems is \emph{significantly} different). When observing outcomes of statistical tests over human judgement and over automatic metric, we get four possible outcomes:

\begin{table}[H]
\centering
\begin{tabular}{cc|c|c}
& &  \multicolumn{2}{c}{Statistical test on a metric} \\
 & & Signif. & Not signif. \\
\hline
\rotmulticell{2}{Humans}
& Signif. & \multicell{Truly differing \\ system pair} & \multicell{Type II \\ Error} \\
\cline{2-4}
& \multicell{Not \\signif.}      & \multicell{Type I. \\ Error} & \multicell{Systems with the \\ equal MT Quality} \\
\end{tabular}
\end{table}

There are two outcomes for the statistical test over a metric that we investigate separately.

In the first scenario, the bootstrap resampling confirms the statistical difference between systems. However, even when both tests agree that systems have statistically different MT quality, it still may happen that humans and metrics disagree on which system is better than the other. The goal is to evaluate how accurate metric decisions are if we employ statistical testing. Therefore, we are interested in the accuracy of a metric over system pairs that are deemed statistically different according to the paired bootstrap resampling, in other words, accuracy for system pairs that are either truly different (top left quadrant) or fall into type I. error (bottom left quadrant).

In the second scenario, we want to find out how many system pairs are diagnosed as non-significant even though human judgements would deem them different. For this scenario, we investigate for how many system pairs bootstrap resampling fails to reject the null hypothesis. However, keep in mind that two statistical tests cannot be directly compared because different tests have different power and the type II error will differ based on that.

\begin{table*}
\centering
\scalebox{0.9}{
\begin{tabular}{lrr|lrr|lrr}
\toprule
Language pair &  Sys. &  Size &         Language pair &  Sys. &  Size &         Language pair &  Sys. &  Size \\
\midrule
     English - French &   145 &  1034 &       English - Hindi &    58 &   540 &   English - Ukrainian &    25 &   988 \\
     English - German &   139 &  2544 &      Polish - English &    57 &  1229 &      English - Slovak &    25 &  1776 \\
     French - English &   131 &  1119 &  Portuguese - English &    57 &   878 &       English - Irish &    24 &   463 \\
     German - English &   122 &  1212 &     Swedish - English &    57 &  1116 &     English - Persian &    24 &   510 \\
   Japanese - English &    78 &   925 &      English - Arabic &    56 &  1054 &      Slovak - English &    23 &  1476 \\
    Chinese - English &    74 &  1029 &      Korean - English &    56 &  1462 &       Greek - English &    23 &  1526 \\
    Italian - English &    71 &  1156 &       Czech - English &    55 &  1105 &    English - Croatian &    22 &  1625 \\
 English - Portuguese &    70 &  1679 &   English - Hungarian &    55 &  1018 &       English - Welsh &    22 &   497 \\
   English - Japanese &    67 &   998 &      English - Korean &    55 &   550 &   English - Norwegian &    22 &  1533 \\
    English - Swedish &    66 &  1219 &     English - Turkish &    55 &  1043 &      English - Hebrew &    22 &   940 \\
    English - Chinese &    65 &  2443 &        English - Thai &    54 &   510 &  English - Vietnamese &    20 &  1857 \\
     English - Danish &    64 &  1186 &       Hindi - English &    54 &   816 &       Welsh - English &    20 &  1686 \\
    English - Italian &    64 &  1505 &     Turkish - English &    54 &  1037 &  Vietnamese - English &    20 &  1697 \\
     English - Polish &    64 &  1188 &      Danish - English &    52 &   986 &     Catalan - English &    20 &   928 \\
    Spanish - English &    64 &  1223 &     English - Russian &    49 &  1159 &        English - Urdu &    18 &   448 \\
      Dutch - English &    63 &   927 &     Russian - English &    44 &   736 &     English - Finnish &    17 &  1802 \\
      English - Dutch &    61 &   991 &        Thai - English &    39 &   457 &       Tamil - English &    16 &   834 \\
 English - Indonesian &    61 &   948 &     English - Catalan &    30 &   981 &  English - Lithuanian &    16 &  1997 \\
 Indonesian - English &    60 &   703 &      Hebrew - English &    28 &   870 &  Lithuanian - English &    16 &  1997 \\
      English - Czech &    59 &  1329 &    English - Romanian &    27 &  1056 &     English - Maltese &    16 &   489 \\
     Arabic - English &    59 &  2674 &    Romanian - English &    27 &  1094 &   English - Kiswahili &    16 &   457 \\
    English - Spanish &    58 &  1172 &       English - Greek &    27 &  1936 &                       &       &       \\
  Hungarian - English &    58 &   976 &     Persian - English &    26 &  1372 &                       &       &       \\
\bottomrule
\end{tabular}
}
\caption{The column ``Sys.'' represents the number of systems for a given translation direction. We list only translation directions with more than \countoflanguagesShowPairsOver{} evaluated systems. The column ``Size'' represents the average test set size for the given direction. We evaluate \countofpairs{} translation directions in total.}
\label{tab:lang_pairs_counts}
\end{table*}

\begin{table}[t]
\centering
\scalebox{0.85}{
\begin{tabular}{lrrrrr}
\toprule
{} &                                All &                               0.05 &                             Within &      Spearman &       Pearson \\
n            &                               3344 &                               1717 &                                541 &          3347 &          3347 \\
\midrule
COMET        &  \cellcolor{black!15}\textbf{83.4} &  \cellcolor{black!15}\textbf{96.5} &  \cellcolor{black!15}\textbf{90.6} &  \textbf{0.879} &  \textbf{0.919} \\
COMET-src    &           \cellcolor{black!15}83.2 &                               95.3 &           \cellcolor{black!15}89.1 &           0.824 &           0.855 \\
Prism        &                               80.6 &                               94.5 &                               86.3 &           0.827 &           0.839 \\
BLEURT-large &                               80.1 &                               94.4 &                               85.4 &           0.808 &           0.748 \\
BLEURT       &                               80.0 &                               93.8 &                               84.1 &           0.787 &           0.729 \\
ESIM         &                               78.7 &                               92.9 &                               82.8 &           0.780 &           0.835 \\
BERTScore    &                               78.3 &                               92.2 &                               81.0 &           0.772 &           0.824 \\
ChrF         &                               75.6 &                               89.5 &                               75.0 &           0.716 &           0.739 \\
TER          &                               75.6 &                               89.2 &                               73.9 &           0.708 &           0.321 \\
CharacTER    &                               74.9 &                               88.6 &                               74.1 &           0.700 &           0.757 \\
BLEU         &                               74.6 &                               88.2 &                               74.3 &           0.661 &           0.640 \\
Prism-src    &                               73.4 &                               85.3 &                               77.4 &           0.661 &           0.631 \\
EED          &                               68.8 &                               79.4 &                               68.2 &           0.531 &           0.541 \\
\bottomrule
\end{tabular}
}
\caption{Extended \cref{tab:system_level_description} with Spearman's and Pearson's correlations over all system pairs. Remaining columns are identical to original table. This table also contain additional BLEURT-large.}
\label{tab:system_level_description_appendix}
\end{table}

\begin{figure*}
\begin{center}
    \includegraphics{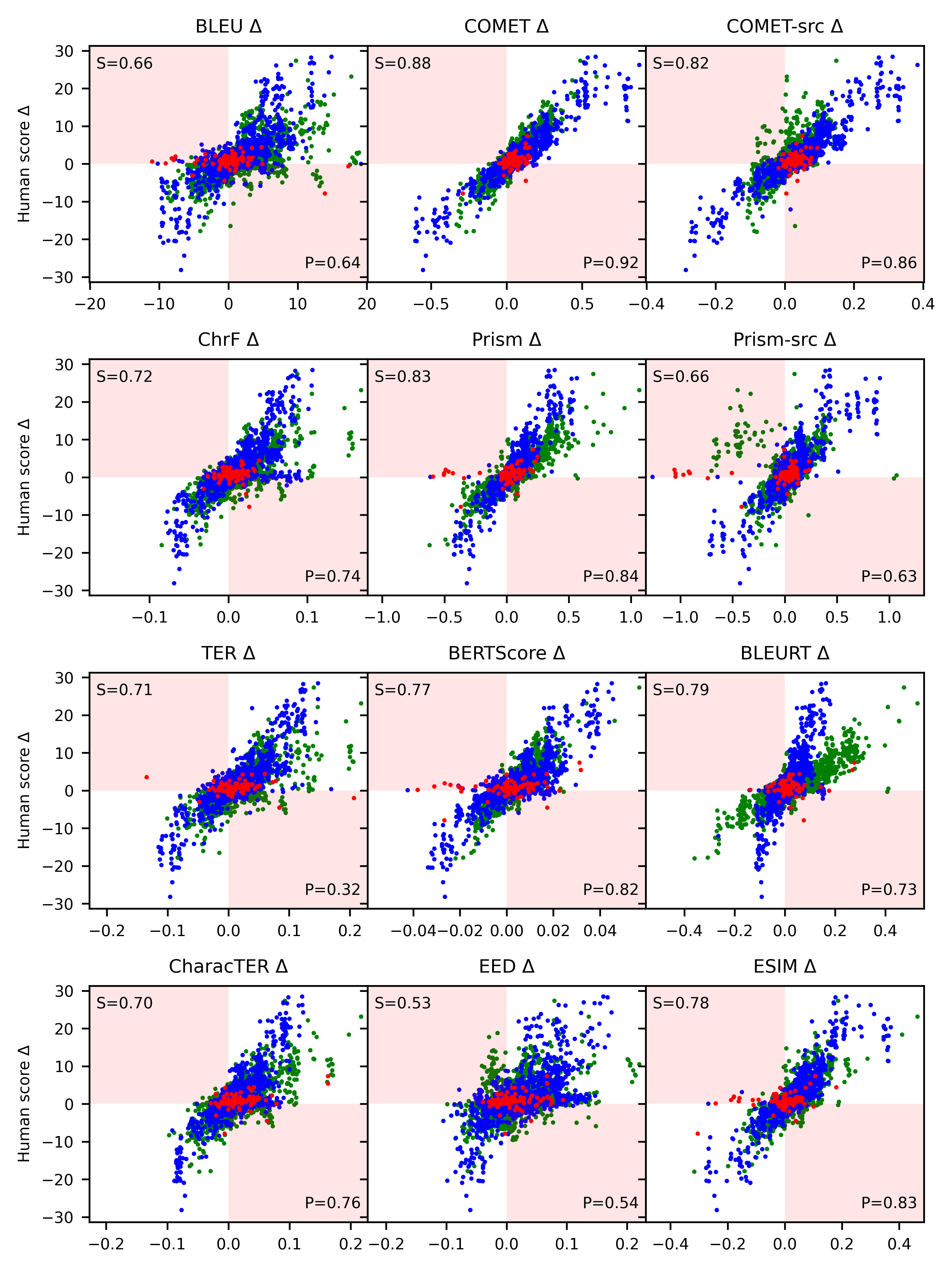}
\end{center}
\caption{\XXXX{switch to SVG}Each point represents a difference in average human judgement (y-axis) and a difference in automatic metric (x-axis) over a pair of systems. Blue points are system pairs translating from English; green points are into English; red points are non-English systems (French, German, and Chinese centric). Spearman's \textrho{} correlation is in top left corner, while Pearson's r is in the bottom right corner. Metrics disagree with human ranking for system pairs in pink quadrants. For better visualization, we have clipped few outliers in BLEU, ChrF, and TER plots.}
\label{app:metric_deltas}
\end{figure*}

\begin{sidewaystable*}
\centering
\begin{tabular}{llllllllllllll}
\toprule
{} &   n &                               COMET &                           COMET-src &                              BLEURT &                               Prism &                           Prism-src &                                ESIM &                                BLEU &                                ChrF &                           BERTScore &                           CharacTER &                                 TER &                                 EED \\
\midrule
English - French     &  62 &   \cellcolor{black!15}\textbf{98.4} &                                93.5 &                                90.3 &                                88.7 &                                87.1 &                                82.3 &                                64.5 &                                64.5 &                                62.9 &                                62.9 &                                58.1 &                                41.9 \\
Japanese - English   &  58 &   \cellcolor{black!15}\textbf{98.3} &                                89.7 &                                84.5 &   \cellcolor{black!15}\textbf{98.3} &                                89.7 &   \cellcolor{black!15}\textbf{98.3} &                                93.1 &                                93.1 &            \cellcolor{black!15}96.6 &                                87.9 &            \cellcolor{black!15}96.6 &                                82.8 \\
English - Polish     &  57 &  \cellcolor{black!15}\textbf{100.0} &  \cellcolor{black!15}\textbf{100.0} &            \cellcolor{black!15}98.2 &  \cellcolor{black!15}\textbf{100.0} &                                82.5 &  \cellcolor{black!15}\textbf{100.0} &  \cellcolor{black!15}\textbf{100.0} &  \cellcolor{black!15}\textbf{100.0} &  \cellcolor{black!15}\textbf{100.0} &  \cellcolor{black!15}\textbf{100.0} &  \cellcolor{black!15}\textbf{100.0} &            \cellcolor{black!15}98.2 \\
Polish - English     &  55 &                                80.0 &   \cellcolor{black!15}\textbf{98.2} &                                83.6 &                                83.6 &   \cellcolor{black!15}\textbf{98.2} &                                83.6 &                                83.6 &                                80.0 &                                83.6 &                                80.0 &                                81.8 &                                81.8 \\
Thai - English       &  54 &  \cellcolor{black!15}\textbf{100.0} &                                70.4 &            \cellcolor{black!15}96.3 &                                90.7 &                                16.7 &                                88.9 &                                83.3 &                                92.6 &                                90.7 &                                94.4 &                                83.3 &                                20.4 \\
English - German     &  52 &  \cellcolor{black!15}\textbf{100.0} &  \cellcolor{black!15}\textbf{100.0} &                                80.8 &  \cellcolor{black!15}\textbf{100.0} &                                88.5 &  \cellcolor{black!15}\textbf{100.0} &                                75.0 &                                67.3 &                                82.7 &                                73.1 &                                78.8 &                                75.0 \\
Chinese - English    &  52 &            \cellcolor{black!15}96.2 &                                94.2 &            \cellcolor{black!15}98.1 &                                92.3 &                                71.2 &  \cellcolor{black!15}\textbf{100.0} &                                94.2 &            \cellcolor{black!15}96.2 &                                94.2 &            \cellcolor{black!15}96.2 &                                86.5 &                                71.2 \\
Turkish - English    &  51 &            \cellcolor{black!15}98.0 &  \cellcolor{black!15}\textbf{100.0} &  \cellcolor{black!15}\textbf{100.0} &            \cellcolor{black!15}96.1 &  \cellcolor{black!15}\textbf{100.0} &            \cellcolor{black!15}96.1 &            \cellcolor{black!15}96.1 &            \cellcolor{black!15}96.1 &            \cellcolor{black!15}96.1 &            \cellcolor{black!15}98.0 &            \cellcolor{black!15}96.1 &                                94.1 \\
English - Indonesian &  51 &  \cellcolor{black!15}\textbf{100.0} &  \cellcolor{black!15}\textbf{100.0} &            \cellcolor{black!15}96.1 &  \cellcolor{black!15}\textbf{100.0} &                                88.2 &  \cellcolor{black!15}\textbf{100.0} &                                92.2 &            \cellcolor{black!15}98.0 &  \cellcolor{black!15}\textbf{100.0} &            \cellcolor{black!15}96.1 &            \cellcolor{black!15}98.0 &            \cellcolor{black!15}96.1 \\
English - Turkish    &  48 &  \cellcolor{black!15}\textbf{100.0} &  \cellcolor{black!15}\textbf{100.0} &  \cellcolor{black!15}\textbf{100.0} &  \cellcolor{black!15}\textbf{100.0} &  \cellcolor{black!15}\textbf{100.0} &  \cellcolor{black!15}\textbf{100.0} &                                87.5 &  \cellcolor{black!15}\textbf{100.0} &  \cellcolor{black!15}\textbf{100.0} &  \cellcolor{black!15}\textbf{100.0} &            \cellcolor{black!15}95.8 &                                91.7 \\
English - Swedish    &  48 &                                91.7 &                                91.7 &   \cellcolor{black!15}\textbf{97.9} &   \cellcolor{black!15}\textbf{97.9} &   \cellcolor{black!15}\textbf{97.9} &            \cellcolor{black!15}95.8 &            \cellcolor{black!15}93.8 &                                87.5 &            \cellcolor{black!15}95.8 &            \cellcolor{black!15}95.8 &            \cellcolor{black!15}93.8 &            \cellcolor{black!15}91.7 \\
Swedish - English    &  47 &            \cellcolor{black!15}97.9 &  \cellcolor{black!15}\textbf{100.0} &            \cellcolor{black!15}97.9 &            \cellcolor{black!15}95.7 &                                89.4 &                                91.5 &                                87.2 &                                91.5 &            \cellcolor{black!15}95.7 &                                72.3 &                                91.5 &                                85.1 \\
English - Dutch      &  47 &            \cellcolor{black!15}97.9 &            \cellcolor{black!15}97.9 &  \cellcolor{black!15}\textbf{100.0} &  \cellcolor{black!15}\textbf{100.0} &            \cellcolor{black!15}97.9 &  \cellcolor{black!15}\textbf{100.0} &  \cellcolor{black!15}\textbf{100.0} &  \cellcolor{black!15}\textbf{100.0} &  \cellcolor{black!15}\textbf{100.0} &  \cellcolor{black!15}\textbf{100.0} &  \cellcolor{black!15}\textbf{100.0} &                                87.2 \\
Indonesian - English &  46 &            \cellcolor{black!15}97.8 &  \cellcolor{black!15}\textbf{100.0} &            \cellcolor{black!15}97.8 &            \cellcolor{black!15}97.8 &  \cellcolor{black!15}\textbf{100.0} &            \cellcolor{black!15}97.8 &                                93.5 &            \cellcolor{black!15}97.8 &            \cellcolor{black!15}97.8 &            \cellcolor{black!15}97.8 &                                91.3 &                                89.1 \\
English - Hungarian  &  46 &  \cellcolor{black!15}\textbf{100.0} &  \cellcolor{black!15}\textbf{100.0} &            \cellcolor{black!15}97.8 &  \cellcolor{black!15}\textbf{100.0} &  \cellcolor{black!15}\textbf{100.0} &            \cellcolor{black!15}97.8 &                                84.8 &                                82.6 &  \cellcolor{black!15}\textbf{100.0} &  \cellcolor{black!15}\textbf{100.0} &  \cellcolor{black!15}\textbf{100.0} &                                87.0 \\
Czech - English      &  44 &  \cellcolor{black!15}\textbf{100.0} &                                93.2 &  \cellcolor{black!15}\textbf{100.0} &  \cellcolor{black!15}\textbf{100.0} &                                93.2 &  \cellcolor{black!15}\textbf{100.0} &  \cellcolor{black!15}\textbf{100.0} &  \cellcolor{black!15}\textbf{100.0} &            \cellcolor{black!15}95.5 &            \cellcolor{black!15}95.5 &            \cellcolor{black!15}95.5 &                                20.5 \\
English - Czech      &  42 &  \cellcolor{black!15}\textbf{100.0} &  \cellcolor{black!15}\textbf{100.0} &                                92.9 &            \cellcolor{black!15}97.6 &            \cellcolor{black!15}97.6 &            \cellcolor{black!15}97.6 &                                92.9 &                                92.9 &            \cellcolor{black!15}97.6 &                                92.9 &            \cellcolor{black!15}97.6 &  \cellcolor{black!15}\textbf{100.0} \\
English - Danish     &  42 &            \cellcolor{black!15}95.2 &            \cellcolor{black!15}95.2 &            \cellcolor{black!15}88.1 &   \cellcolor{black!15}\textbf{97.6} &            \cellcolor{black!15}95.2 &            \cellcolor{black!15}92.9 &   \cellcolor{black!15}\textbf{97.6} &            \cellcolor{black!15}92.9 &            \cellcolor{black!15}95.2 &            \cellcolor{black!15}92.9 &            \cellcolor{black!15}95.2 &                                88.1 \\
Hungarian - English  &  42 &            \cellcolor{black!15}90.5 &                                83.3 &                                78.6 &   \cellcolor{black!15}\textbf{97.6} &                                85.7 &                                90.5 &            \cellcolor{black!15}90.5 &            \cellcolor{black!15}95.2 &                                85.7 &            \cellcolor{black!15}92.9 &                                81.0 &            \cellcolor{black!15}88.1 \\
Hindi - English      &  41 &            \cellcolor{black!15}95.1 &  \cellcolor{black!15}\textbf{100.0} &            \cellcolor{black!15}95.1 &                                82.9 &                                73.2 &                                78.0 &                                61.0 &                                61.0 &                                80.5 &                                65.9 &                                73.2 &                                80.5 \\
Dutch - English      &  40 &  \cellcolor{black!15}\textbf{100.0} &            \cellcolor{black!15}97.5 &            \cellcolor{black!15}97.5 &            \cellcolor{black!15}97.5 &                                82.5 &            \cellcolor{black!15}97.5 &            \cellcolor{black!15}97.5 &            \cellcolor{black!15}95.0 &            \cellcolor{black!15}97.5 &            \cellcolor{black!15}97.5 &            \cellcolor{black!15}97.5 &                                75.0 \\
German - English     &  39 &   \cellcolor{black!15}\textbf{97.4} &            \cellcolor{black!15}94.9 &   \cellcolor{black!15}\textbf{97.4} &   \cellcolor{black!15}\textbf{97.4} &                                82.1 &            \cellcolor{black!15}94.9 &                                82.1 &                                82.1 &   \cellcolor{black!15}\textbf{97.4} &                                82.1 &            \cellcolor{black!15}94.9 &                                64.1 \\
Danish - English     &  39 &  \cellcolor{black!15}\textbf{100.0} &  \cellcolor{black!15}\textbf{100.0} &  \cellcolor{black!15}\textbf{100.0} &  \cellcolor{black!15}\textbf{100.0} &  \cellcolor{black!15}\textbf{100.0} &  \cellcolor{black!15}\textbf{100.0} &            \cellcolor{black!15}97.4 &            \cellcolor{black!15}97.4 &  \cellcolor{black!15}\textbf{100.0} &                                89.7 &            \cellcolor{black!15}97.4 &                                74.4 \\
Spanish - English    &  39 &            \cellcolor{black!15}94.9 &            \cellcolor{black!15}94.9 &            \cellcolor{black!15}94.9 &   \cellcolor{black!15}\textbf{97.4} &            \cellcolor{black!15}92.3 &            \cellcolor{black!15}94.9 &            \cellcolor{black!15}94.9 &            \cellcolor{black!15}94.9 &   \cellcolor{black!15}\textbf{97.4} &   \cellcolor{black!15}\textbf{97.4} &            \cellcolor{black!15}94.9 &            \cellcolor{black!15}92.3 \\
English - Russian    &  37 &  \cellcolor{black!15}\textbf{100.0} &  \cellcolor{black!15}\textbf{100.0} &  \cellcolor{black!15}\textbf{100.0} &  \cellcolor{black!15}\textbf{100.0} &                                91.9 &  \cellcolor{black!15}\textbf{100.0} &  \cellcolor{black!15}\textbf{100.0} &  \cellcolor{black!15}\textbf{100.0} &  \cellcolor{black!15}\textbf{100.0} &  \cellcolor{black!15}\textbf{100.0} &  \cellcolor{black!15}\textbf{100.0} &  \cellcolor{black!15}\textbf{100.0} \\
Portuguese - English &  36 &  \cellcolor{black!15}\textbf{100.0} &            \cellcolor{black!15}94.4 &            \cellcolor{black!15}94.4 &            \cellcolor{black!15}94.4 &            \cellcolor{black!15}94.4 &            \cellcolor{black!15}94.4 &            \cellcolor{black!15}94.4 &                                88.9 &            \cellcolor{black!15}94.4 &            \cellcolor{black!15}94.4 &            \cellcolor{black!15}94.4 &            \cellcolor{black!15}97.2 \\
Korean - English     &  33 &  \cellcolor{black!15}\textbf{100.0} &            \cellcolor{black!15}97.0 &                                72.7 &                                87.9 &                                42.4 &                                69.7 &                                63.6 &            \cellcolor{black!15}97.0 &                                78.8 &                                90.9 &                                69.7 &                                66.7 \\
English - Portuguese &  31 &  \cellcolor{black!15}\textbf{100.0} &  \cellcolor{black!15}\textbf{100.0} &  \cellcolor{black!15}\textbf{100.0} &  \cellcolor{black!15}\textbf{100.0} &                                87.1 &  \cellcolor{black!15}\textbf{100.0} &            \cellcolor{black!15}96.8 &  \cellcolor{black!15}\textbf{100.0} &  \cellcolor{black!15}\textbf{100.0} &  \cellcolor{black!15}\textbf{100.0} &            \cellcolor{black!15}96.8 &  \cellcolor{black!15}\textbf{100.0} \\
English - Italian    &  28 &  \cellcolor{black!15}\textbf{100.0} &  \cellcolor{black!15}\textbf{100.0} &  \cellcolor{black!15}\textbf{100.0} &  \cellcolor{black!15}\textbf{100.0} &            \cellcolor{black!15}96.4 &  \cellcolor{black!15}\textbf{100.0} &            \cellcolor{black!15}96.4 &            \cellcolor{black!15}96.4 &  \cellcolor{black!15}\textbf{100.0} &  \cellcolor{black!15}\textbf{100.0} &            \cellcolor{black!15}96.4 &                                75.0 \\
English - Japanese   &  28 &            \cellcolor{black!15}96.4 &                                89.3 &  \cellcolor{black!15}\textbf{100.0} &            \cellcolor{black!15}96.4 &                                75.0 &            \cellcolor{black!15}96.4 &                                78.6 &            \cellcolor{black!15}96.4 &            \cellcolor{black!15}96.4 &                                67.9 &                                78.6 &                                46.4 \\
Russian - English    &  27 &            \cellcolor{black!15}92.6 &  \cellcolor{black!15}\textbf{100.0} &  \cellcolor{black!15}\textbf{100.0} &                                88.9 &                                85.2 &                                81.5 &                                66.7 &                                59.3 &                                88.9 &                                40.7 &                                74.1 &                                59.3 \\
English - Spanish    &  25 &            \cellcolor{black!15}92.0 &            \cellcolor{black!15}88.0 &            \cellcolor{black!15}88.0 &   \cellcolor{black!15}\textbf{96.0} &                                76.0 &            \cellcolor{black!15}92.0 &            \cellcolor{black!15}92.0 &            \cellcolor{black!15}88.0 &            \cellcolor{black!15}92.0 &                                84.0 &                                72.0 &                                84.0 \\
\bottomrule
\end{tabular}
\caption{Each row represents accuracy of system pairs for given language pair. We list language pairs with at least 20 system pairs. Results are calculated over a set of significantly different system pairs with alpha level 0.05. Results with grey background are considered to be tied with the best metric. Interestingly, when we investigated Polish--English results, we found out the test set is likely post-edited MT output.}
\label{tab:individual_language_pairs}
\end{sidewaystable*}

\end{document}